\pdfoutput=1
\documentclass{article}

\usepackage[preprint]{neurips_2026}

\usepackage[utf8]{inputenc}
\usepackage[T1]{fontenc}
\usepackage{hyperref}
\usepackage{url}
\usepackage{booktabs}
\usepackage{amsfonts,amsmath,amssymb,amsthm}
\usepackage{nicefrac}
\usepackage{microtype}
\usepackage{xcolor}
\usepackage{graphicx}
\usepackage{algorithm}
\usepackage{algorithmic}
\usepackage{multirow}
\usepackage{pifont}
\newcommand{\cmark}{\ding{51}}
\newcommand{\xmark}{\ding{55}}
\usepackage[most]{tcolorbox}
\newtcolorbox{promptbox}[1]{enhanced, colback=black!3, colframe=black!35, boxrule=0.4pt, arc=1.5pt,
  left=4pt, right=4pt, top=3pt, bottom=3pt, title={#1}, fonttitle=\small\bfseries,
  coltitle=black, colbacktitle=black!8, attach boxed title to top left={yshift=-2mm, xshift=3mm},
  boxed title style={boxrule=0.4pt, arc=1pt}, breakable}

\newcommand{\st}[2][t]{s_{#1}(#2)}      %
\newcommand{\mt}[1][t]{m_{#1}}           %
\newcommand{\Tstop}{T_{\mathrm{stop}}} %
\newcommand{\keeprate}{\kappa}         %
\newcommand{\rse}{\mathrm{RSE}}        %
\newcommand{\method}{ProbeGuard}  %
\newcommand{\repourl}{\url{https://github.com/wangxiaoyang0412/probeguard}}  %

\newcommand{\UaRateMedQA}{59}
\newcommand{\UaNMedQA}{746}
\newcommand{\UaErrNMedQA}{100}
\newcommand{\UaErrRateMedQA}{13.4}
\newcommand{\UaRateMMLUPro}{59}
\newcommand{\UaErrRateMMLUPro}{17.7}
\newcommand{\UaRateMedXpert}{29}
\newcommand{\UaErrRateMedXpert}{79.0}
\newcommand{\UaRateMedMCQA}{55}
\newcommand{\UaErrRateMedMCQA}{22.3}
\newcommand{\UaErrRateMedQAHuatuo}{6.2}
\newcommand{\CombinedMedQA}{0.696}
\newcommand{\TerminalMedQA}{0.512}
\newcommand{\ROneMedQA}{0.677}
\newcommand{\SeqLogprobMedQA}{0.400}
\newcommand{\AnsEntropyMedQA}{0.681}
\newcommand{\CombinedMMLUPro}{0.704}
\newcommand{\TerminalMMLUPro}{0.504}
\newcommand{\ROneMMLUPro}{0.698}
\newcommand{\SeqLogprobMMLUPro}{0.474}
\newcommand{\AnsEntropyMMLUPro}{0.703}
\newcommand{\CombinedMedMCQA}{0.659}
\newcommand{\TerminalMedMCQA}{0.507}
\newcommand{\ROneMedMCQA}{0.640}
\newcommand{\SeqLogprobMedMCQA}{0.534}
\newcommand{\AnsEntropyMedMCQA}{0.641}
\newcommand{\CombinedMedQAHuatuo}{0.775}
\newcommand{\TerminalMedQAHuatuo}{0.519}
\newcommand{\CombinedMMLUProHuatuo}{0.701}
\newcommand{\TerminalMMLUProHuatuo}{0.526}
\newcommand{\CombinedMedQALlama}{0.825}
\newcommand{\TerminalMedQALlama}{0.548}
\newcommand{\CombinedMMLUProLlama}{0.782}
\newcommand{\TerminalMMLUProLlama}{0.551}
\newcommand{\AurcTerminalMedQA}{0.221}
\newcommand{\AurcROneMedQA}{0.155}
\newcommand{\AurcSeqLogprobMedQA}{0.320}
\newcommand{\AurcAnsEntropyMedQA}{0.154}
\newcommand{\AurcCombinedMedQA}{0.145}
\newcommand{\AurcVerbalizedMedQA}{0.179}
\newcommand{\AurcPTrueMedQA}{0.160}
\newcommand{\AurcTerminalMMLUPro}{0.300}
\newcommand{\AurcROneMMLUPro}{0.200}
\newcommand{\AurcSeqLogprobMMLUPro}{0.376}
\newcommand{\AurcAnsEntropyMMLUPro}{0.199}
\newcommand{\AurcCombinedMMLUPro}{0.205}
\newcommand{\AurcVerbalizedMMLUPro}{0.207}
\newcommand{\AurcPTrueMMLUPro}{0.205}
\newcommand{\AurcTerminalMedMCQA}{0.330}
\newcommand{\AurcROneMedMCQA}{0.250}
\newcommand{\AurcSeqLogprobMedMCQA}{0.347}
\newcommand{\AurcAnsEntropyMedMCQA}{0.250}
\newcommand{\AurcCombinedMedMCQA}{0.234}
\newcommand{\AurcTerminalMedQAHuatuo}{0.193}
\newcommand{\AurcCombinedMedQAHuatuo}{0.096}
\newcommand{\AurcTerminalMMLUProHuatuo}{0.353}
\newcommand{\AurcCombinedMMLUProHuatuo}{0.225}
\newcommand{\AurcTerminalMedQALlama}{0.289}
\newcommand{\AurcCombinedMedQALlama}{0.142}
\newcommand{\AurcTerminalMMLUProLlama}{0.406}
\newcommand{\AurcCombinedMMLUProLlama}{0.246}
\newcommand{\VerbalizedMedQA}{0.585}
\newcommand{\VerbalizedMMLUPro}{0.651}
\newcommand{\PTrueMedQA}{0.602}
\newcommand{\PTrueMMLUPro}{0.644}
\newcommand{\FlipWrongMedMCQA}{26.5}
\newcommand{\FlipRightMedMCQA}{5.5}
\newcommand{\KeepMedMCQA}{0.729}
\newcommand{\FlipRightMMLUPro}{3.3}
\newcommand{\KeepMMLUPro}{0.638}
\newcommand{\FlipWrongMedQAHuatuo}{15.4}
\newcommand{\FlipRightMedQAHuatuo}{3.7}
\newcommand{\FlipWrongMedQALlama}{54.2}
\newcommand{\FlipRightMedQALlama}{28.3}
\newcommand{\FlipWrongCtrlNodocs}{0.0}
\newcommand{\FlipRightCtrlNodocs}{0.0}
\newcommand{\FlipWrongCtrlRanddocs}{1.0}
\newcommand{\FlipRightCtrlRanddocs}{0.2}
\newcommand{\FlipWrongCtrlNeutral}{12.0}
\newcommand{\FlipRightCtrlNeutral}{3.6}
\newcommand{\NaiveViolation}{15.7}
\newcommand{\CertGlobalShare}{32}
\newcommand{\AblDropFThree}{0.682}
\newcommand{\HardFThreeMedQA}{0.578}
\newcommand{\HardFThreeMMLUPro}{0.593}
\newcommand{\RhoTrajMin}{0.93}
\newcommand{\RhoRseMax}{0.39}
\newcommand{\SmokeDelta}{0.007}
\newcommand{\FThreeMedQA}{0.587}
\newcommand{\FThreeMMLUPro}{0.613}
\newcommand{\KeepMedQA}{0.642}
\newcommand{\ComboUaMedQA}{0.685}
\newcommand{\ComboUaMMLUPro}{0.688}
\newcommand{\SlpUaMedQA}{0.680}
\newcommand{\PTrueUaMedQA}{0.602}
\newcommand{\VerbUaMedQA}{0.599}
\newcommand{\ComboThreeUaMedQA}{0.730}
\newcommand{\FlipWrongMedQA}{14.0}
\newcommand{\FlipRightMedQA}{4.2}
\newcommand{\CtrlDiffNodocsLo}{3.0}
\newcommand{\CtrlDiffNodocsHi}{16.8}
\newcommand{\CtrlDiffRanddocsLo}{1.7}
\newcommand{\CtrlDiffRanddocsHi}{16.8}
\newcommand{\CertAlpha}{0.15}
\newcommand{\CertCovMean}{0.553}
\newcommand{\CertCovSE}{0.038}
\newcommand{\CertRisk}{9.0}
\newcommand{\CertRepeats}{50}
\newcommand{\CertViolations}{0}
\newcommand{\FuelN}{875}
\newcommand{\FuelCertCovTen}{0.219}
\newcommand{\FuelCertCovTenSE}{0.023}
\newcommand{\FuelCertRiskTen}{6.5}
\newcommand{\FuelCertCovFifteen}{0.892}
\newcommand{\FuelCertCovFifteenSE}{0.004}
\newcommand{\CertCovTenHuatuo}{0.410}
\newcommand{\CertCovFifteenHuatuo}{0.997}
\newcommand{\CertCovFifteenLlama}{0.838}
\newcommand{\FormationRFourErr}{58.2}
\newcommand{\ProbeMedianSecMedQA}{114}
\newcommand{\ProbeMedianTok}{4381}
\newcommand{\GenMedianSecMedQA}{72}

\newtheorem{theorem}{Theorem}
\newtheorem{assumption}{Assumption}

\title{Unanimously Wrong: Certified Abstention from How Medical LLM Consensus Forms}

\author{%
  Xiaoyang Wang \\
  Drexel University \\
  \texttt{xw388@drexel.edu} \\
  \And
  Tianrui Wang \\
  University of Michigan, Ann Arbor \\
  \texttt{tianruiw@umich.edu} \\
  \And
  Christopher C. Yang \\
  Drexel University \\
  \texttt{chris.yang@drexel.edu} \\
}

\begin{document}

\maketitle

\begin{abstract}
In clinical practice, agreement among independent experts is treated as evidence of reliability, and multi-round consensus has become a core mechanism of agentic medical question-answering systems. When such a system must decide whether to trust its own answer, the prevailing signal is again agreement, now among the sampled answers. But agreement is a fragile proxy for correctness. A system can be \emph{unanimously wrong}, returning the same incorrect answer on every sample, and on these questions agreement-based signals carry no information. The cause is that these signals read only the final state of the consensus and discard how it was reached. Agreement that was reached by resolving disagreement with evidence looks identical, at the end, to agreement that was present from the first sample because every sample shares one misconception. \textbf{\method{}} is a certified abstention framework that bases the abstention decision on how the consensus formed. All of its components are computed from the system's execution log. Process features trace agreement trajectories, minority persistence, and retrieval saturation. For unanimous votes, \emph{rationale semantic entropy} checks whether the reasons behind the vote cohere, and an active probe retrieves counter-evidence and measures whether the consensus survives. A stratified Learn-then-Test calibration then converts these scores into a distribution-free bound on selective risk. This lets the system separate unanimous votes that are backed by knowledge from those that are not, and answer only where its error rate is provably controlled. We evaluate \method{} on three medical QA benchmarks and a hard-frontier reference, with a published multi-round agentic RAG substrate, against six abstention baselines. On MedQA, \UaErrRateMedQA\% of unanimous votes are wrong, and no agreement-based signal can flag them. Process signals raise the discrimination of correct from incorrect consensus from chance to \CombinedMedQA{} AUROC. The certified rule answers six in ten unanimous-layer questions at an observed selective risk of \CertRisk\%, and nine in ten once in-domain calibration data accumulate. Code: \repourl.

\end{abstract}

\section{Introduction}
\label{sec:intro}

\begin{figure}[t]
  \centering
  \includegraphics[width=0.88\linewidth]{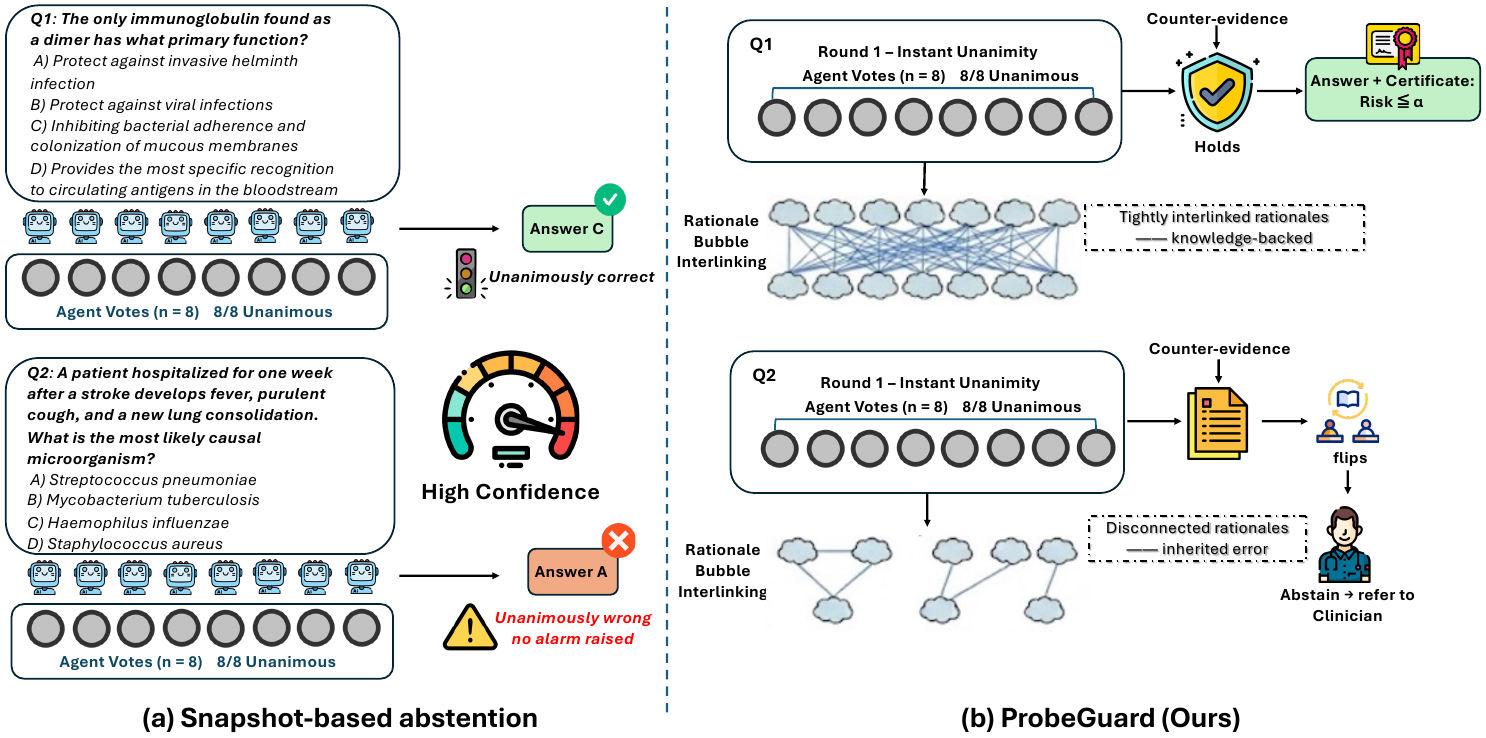}
  \caption{Two unanimous consensuses are indistinguishable to snapshot-based abstention (a), yet one is knowledge-backed and one is inherited from a shared misconception. \method{} (b) checks whether the rationales behind a unanimous vote support one another and stress-tests the consensus with retrieved counter-evidence, answering only where selective risk is certifiably bounded.}
  \label{fig:motivation}
\end{figure}

Language models now encode substantial clinical knowledge \citep{singhal2023clinical} and are moving into clinical question answering, where their answers increasingly inform triage, guideline lookup, and decision support. In these settings the failure that matters most is a \emph{confidently} wrong answer. In a randomized trial, physicians exposed to erroneous LLM suggestions lost 14 percentage points of diagnostic accuracy, despite prior AI-literacy training and voluntary consultation \citep{qazi2026automation}; an answer delivered without any signal of doubt propagates directly into downstream decisions. The model itself cannot be trusted to raise that signal: frontier models almost never abstain when they should, and reasoning-oriented training makes abstention \emph{worse} \citep{kirichenko2025abstentionbench}. A deployed system therefore needs reliable abstention, so that the questions it cannot answer are routed to a clinician \citep{machcha2026medabstain, li2024mediq}.

The dominant abstention signals estimate confidence from a single round of samples: the model's verbalized certainty \citep{kadavath2022know, tian2023justask, xiong2024verbalized}, or the agreement among independently sampled answers \citep{wang2023selfconsistency, manakul2023selfcheckgpt, kuhn2023semantic}. Agreement scores have even been calibrated into conformal abstention rules directly \citep{yadkori2024conformal}. Moving these signals to a modern multi-round consensus system \citep{wu2026marag} exposes their limits. Agreement can be \emph{inherited} from a shared training-data error, in which case a model gives the same wrong answer eight times. Multi-round systems stop deliberating as soon as agreement is reached, so the terminal vote is 1 for every converged question. And on the questions where the system is unanimously wrong, no vote-based signal can raise an alarm. On our substrate, \UaRateMedQA\% of MedQA questions reach first-round unanimity, and \UaErrRateMedQA\% of those unanimous votes are wrong; on expert-level questions the rate rises to \UaErrRateMedXpert\% (Figure~\ref{fig:motivation}).

These signals discard the \emph{formation process}. Consider a panel of physicians who end up agreeing. All of them may have known the answer from the start, or all of them may share one misconception. A forced vote can also produce agreement, and so can a discussion in which they disagreed until one of them found the evidence that settled the question. The final tally is identical in each case, and the differences lie only in the record of the discussion. Multi-round consensus systems write this record to their execution logs, including which candidates disagreed, for how many rounds, which documents were retrieved, and whether the vote changed afterward. Yet prior work discards the trajectory once the answer is produced \citep{wu2026marag, manczak2025shallow}. When the vote is unanimous from the start, the trajectory is trivial and the check moves one level down, to the rationales. The \emph{reasons} behind the vote can be checked for coherence, which extends semantic entropy \citep{farquhar2024semantic} from answers to rationales, and the consensus can be tested directly by supplying opposing evidence.

\textbf{Present work.}
We introduce \method{}, which turns consensus formation into a certified abstention decision. Process features computed from the execution log trace how the vote evolved; for unanimous votes, rationale semantic entropy and an active counter-evidence probe read the evidence behind the vote; and Learn-then-Test calibration \citep{angelopoulos2021ltt, angelopoulos2024crc} within process strata yields a selective-risk bound that holds without distributional assumptions, where existing medical abstention methods provide heuristic scores \citep{dang2026knowguard, shankar2025energy}. Every multi-round system already writes this log, and because the scores are training-free, the entire calibration set can be spent on threshold selection, and the thresholds it returns carry a guarantee a deployment can act on.

We make four contributions. (1) To our knowledge, we give the first abstention decision derived from \emph{how} multi-round consensus forms, with a distribution-free selective-risk guarantee, in place of snapshot-based confidence assessment. (2) We instrument the execution log with trajectory features at zero additional inference cost, extend semantic entropy from answers to the concluding claims of rationales, and introduce an active counter-evidence probe whose \emph{keep rate} measures whether a unanimous vote survives opposing evidence. (3) On questions where all $N{=}8$ candidates agree immediately, every vote-based signal is constant, yet \UaErrNMedQA{} of \UaNMedQA{} such votes on MedQA are wrong. We measure this failure mode across four benchmarks and three backbones. (4) The process score separates correct from incorrect consensus at \CombinedMedQA{} AUROC on MedQA, where terminal agreement scores \TerminalMedQA, and at \CombinedMMLUPro{} on MMLU-Pro medicine. Wrong consensus flips under counter-evidence at \FlipWrongMedQA--\FlipWrongMedMCQA\%, against \FlipRightMMLUPro--\FlipRightMedMCQA\% for correct consensus. Stratified Learn-then-Test calibration answers \CertCovMean{} of the unanimous layer at \CertRisk\% observed risk ($\alpha{=}\CertAlpha$, \CertRepeats{} splits).

\section{Related Work}
\label{sec:related}

\textbf{Abstention in medical question answering.}
Abstention is increasingly treated as a required capability of clinical language systems. KnowGuard grounds abstention in evidence-sufficiency scores over an external medical knowledge graph \citep{dang2026knowguard}; MediQ converts insufficient information into follow-up questions in interactive settings \citep{li2024mediq}; MedAbstain contributes a unified evaluation protocol with explicit abstention options and adversarial variants \citep{machcha2026medabstain}; and, closest to our setting, an energy-based model scores out-of-scope queries for a medical RAG assistant \citep{shankar2025energy}. These systems either require resources beyond the deployed pipeline or produce heuristic scores whose threshold comes with no error guarantee. \method{} needs only the system's own execution log and a probe built from its own retrieval stack, and returns a threshold with a certified selective-risk bound (Table~\ref{tab:resources}).

\begin{table}[t]
\centering
\caption{Comparison of abstention approaches for medical QA. External resources, scoring signals, and risk guarantees are reported for each method.}
\label{tab:resources}
\footnotesize
\setlength{\tabcolsep}{5pt}
\begin{tabular}{lccc}
\toprule
\textbf{Method} & \textbf{External resource} & \textbf{Abstention score} & \textbf{Risk guarantee} \\
\midrule
KnowGuard & medical knowledge graph & evidence sufficiency & \xmark \\
Energy-based scoring & separately trained scorer & energy & \xmark \\
Conformal abstention & none & answer agreement & global \\
\cmidrule(lr){1-4}
\method{} (ours) & none & consensus formation & \textbf{per-stratum} \\
\bottomrule
\end{tabular}
\end{table}

\textbf{Consistency- and confidence-based uncertainty.}
For black-box LLMs, the dominant signal families elicit confidence verbally \citep{kadavath2022know, tian2023justask, xiong2024verbalized} or measure agreement across independently sampled responses: self-consistency \citep{wang2023selfconsistency}, sample-divergence checks \citep{manakul2023selfcheckgpt}, and semantic entropy, which clusters samples by bidirectional entailment \citep{kuhn2023semantic, farquhar2024semantic}; recent surveys place consistency-based estimators among the most widely used black-box approaches \citep{shorinwa2025survey, liu2025kddsurvey}. All of these compute their score from a single round of samples. As discussed in \S\ref{sec:intro}, agreement can be a shared error, and answer-level agreement can hide disagreement at the reasoning level even in debate systems \citep{wang2026illusion}. Our features use the same samples and additionally the rounds between the first sample and the final vote.

\textbf{Multi-round deliberation and its evaluation.}
Agentic RAG pipelines improve question answering by coordinating specialized agents \citep{nguyen2025marag}, and in medical QA by iterating conflict-guided retrieval until consensus \citep{wu2026marag}, while recent evaluations show that reliability degrades under multi-turn interaction \citep{manczak2025shallow} and that reasoning-oriented training can make abstention \emph{worse} \citep{kirichenko2025abstentionbench}. This literature uses deliberation to improve accuracy and leaves the intermediate states unused once the answer is produced. To our knowledge, no prior work uses the deliberation trace as an abstention signal.

\textbf{Conformal guarantees for language models.}
Conformal risk control and Learn-then-Test provide distribution-free selective guarantees \citep{angelopoulos2024crc, angelopoulos2021ltt}, instantiated for LLMs in sampling-based generation \citep{quach2024clm}, factuality filtering \citep{mohri2024factuality}, retrieval-augmented QA \citep{li2024traq}, answer-set prediction under outliers \citep{wang2025sconu}, selective trust of generated reports \citep{gui2024alignment}, and consistency-scored abstention \citep{yadkori2024conformal}; in clinical QA, behavioural features of single reasoning traces have been calibrated with conformal sets \citep{testoni2026mindgap}. In every case the calibrated score is a single-round quantity and the calibration is global over the population. We apply the same procedure to scores computed from the consensus process and calibrate within each process stratum, including the unanimous stratum on which the scores used in prior work are constant.

\section{Method}
\label{sec:method}

\textbf{Problem Formulation.}
Let $x$ be a question posed to a multi-round consensus system. We seek an abstention decision
\begin{equation}
  A : x \;\mapsto\; \{\textsc{answer},\, \textsc{abstain}\},
  \label{eq:decision}
\end{equation}
such that, among the questions the system chooses to answer, the error rate is certifiably bounded by a user-specified level $\alpha$. We build on a published multi-round agentic RAG system for medical QA \citep{wu2026marag}, used unmodified at a pinned commit. At each round $t$ the system samples $N{=}8$ candidate answers together with free-text rationales. If the candidates disagree, it generates conflict-guided queries, retrieves evidence from a medical corpus, injects the evidence into the next round's context, and re-answers, stopping at unanimity or after $T{=}8$ rounds. We add only a logging layer and modify no decision logic. For every round it records the candidate answers with their full rationales, the answer distribution, and all retrieval events (queries and document identifiers). We write $\st{a} \in \{0, \nicefrac{1}{N}, \dots, 1\}$ for the fraction of candidates voting option $a$ at round $t$, $\mt = \arg\max_a \st{a}$ for the modal answer, $\Tstop \le T$ for the stopping round, and $D_t$ for the set of documents retrieved after round $t$. \method{} derives its abstention decision from this log through trajectory features of the consensus process (\S\ref{sec:process}), evidence signals that remain informative when the vote is unanimous (\S\ref{sec:evidence}), and a stratified calibration that certifies a decision rule from these scores (\S\ref{sec:certify}).

\subsection{Consensus-Process Features}
\label{sec:process}

The trajectory recorded in the log carries information that is absent from the terminal vote. Two single-round statistics bracket the trajectory. Terminal agreement $\st[\Tstop]{\mt[\Tstop]}$ equals $1$ for every converged question because of the early-stopping rule. \emph{First-round agreement}
\begin{equation}
  a_1 \;=\; \st[1]{\mt[1]}
  \label{eq:r1}
\end{equation}
is the strongest statistic available to single-round sampling methods. The trajectory between them is summarized by the \emph{trajectory-averaged agreement}, which rewards early, stable consensus over late or forced convergence, and by the \emph{majority-flip count}, the number of rounds at which the leading answer is overturned. Two further families capture who disagrees and what the retriever is doing. \emph{Minority persistence} is the share of rounds in which the alternative answer that persists longest retains a supporter; a single alternative that survives every round signals genuine ambiguity, whereas rotating, short-lived minorities are sampling noise. The \emph{document-repeat rate} is the fraction of newly retrieved documents already seen in earlier rounds; when it rises, the evidence space is saturating and further rounds are unlikely to resolve the disagreement. The full registry comprises 34 features across five families (first-round and trajectory agreement, convergence statistics, minority dynamics, rationale-level signals, and retrieval-side behavior), each with a definition fixed in advance of evaluation; formal definitions are in Appendix~\ref{app:features}. A combined process score is fit by $\ell_2$-regularized logistic regression over these features with question-level cross-validation, median imputation and missingness indicators computed within training folds (Appendix~\ref{app:impl}).

\subsection{Evidence Signals for Unanimous Votes}
\label{sec:evidence}

When the first-round vote is unanimous ($a_1 = 1$), every agreement-based feature in \S\ref{sec:process} is constant: the trajectory has length one and no retrieval is triggered. These are also the questions on which a shared training-data error produces a confident mistake. If the answer rests on knowledge, the eight rationales cite the same mechanism. If it rests on a shared error, no common mechanism exists to cite, and each sample constructs its own justification, so the rationales diverge.

\textbf{Rationale semantic entropy (passive).}
We formalize the question of whether eight identical answers share one justification as \emph{mutual entailment among the concluding claims} of the rationales, treating the consensus as knowledge-backed when its rationales entail one another. This reduces our problem to one with a well-studied estimator, semantic entropy \citep{kuhn2023semantic, farquhar2024semantic}, which clusters sampled \emph{answers} by bidirectional NLI entailment and computes the entropy of the cluster distribution. On a unanimous vote its answer-level value is identically zero, so we apply the estimator to the rationales instead, conditioned on answer agreement. From each of the $N$ rationales we extract the concluding claim by a deterministic rule frozen before any rationale data were inspected (rule and unit tests in Appendix~\ref{app:claims}). We score concluding claims instead of full rationales because full rationales exceed the NLI context length and share enough surface vocabulary that topical similarity masks logical disagreement. Each ordered claim pair is then scored by an NLI model, yielding entailment probabilities $e_{ij}$. Hard entailment clustering saturates at $\ln N$ on the most suspicious questions, so we define the score on the continuous probabilities directly,
\begin{equation}
  \rse \;=\; \operatorname{mean}_{i<j}\, \min(e_{ij},\, e_{ji}),
  \label{eq:rse}
\end{equation}
where the minimum requires each claim pair to support each other in both directions, and the mean runs over all $\binom{N}{2}$ unordered pairs, so that the score rises with the mutual support among the rationales.

\textbf{Counter-evidence probe (active).}
The probe reuses the substrate's retrieval loop with the direction reversed. On contested questions the substrate retrieves evidence to resolve a disagreement; the probe retrieves evidence against an existing agreement and checks whether the agreement survives. If the vote rests on knowledge, the model can state why the challenge does not apply and the vote should survive the strongest available counter-evidence; if it rests on a shared misconception, no mechanism is available to rebut the challenge and we expect the vote to be revised. Concretely, for each unanimous question we (i) prompt the model to generate retrieval queries that would support \emph{alternative} answers or contradict the consensus, (ii) retrieve challenge documents with the system's own retrieval stack under its standard parameters, and (iii) let all $N$ candidates re-answer with the challenge documents in context, under an explicit devil's-advocate instruction that licenses revision without demanding it (prompts in Appendix~\ref{app:prompts}; Algorithm~\ref{alg:probe} in Appendix~\ref{app:algos}). The probe statistic is the \emph{keep rate}
\begin{equation}
  \keeprate \;=\; \tfrac{1}{N}\,\bigl|\{\, i : \text{candidate } i \text{ retains the consensus answer} \,\}\bigr|,
  \label{eq:keep}
\end{equation}
where the count runs over the $N$ re-answers. The unanimous-layer score combines the passive and active signals as $z(\keeprate) + z(\rse)$, where $z(\cdot)$ denotes per-dataset standardization. The probe fires only on the unanimous subset and costs one query-generation call, one retrieval round, and $N$ re-answer calls per probed question; per-question latency and token accounting are reported in Appendix~\ref{app:cost}.

\subsection{Stratified Certification}
\label{sec:certify}

Deployment needs a threshold whose selective risk is bounded with stated confidence, which a point estimate on held-out data does not provide. The unanimous and contested subpopulations differ in base error rate, in which signals are defined, and in score scale, so a single global calibration is dominated by this heterogeneity: thresholds that are valid on average can be badly miscalibrated within each subpopulation. Selective risk is also not monotone in the threshold, since admitting more questions can \emph{lower} the error rate among answered questions, so naive quantile calibration over-rejects or over-admits depending on where the errors sit. We therefore certify per stratum, selecting each threshold with a multiple-testing procedure over a grid of candidates.

The target guarantee is
\begin{equation}
  \Pr\big[\,\mathrm{risk}(\hat{\lambda}) \le \alpha\,\big] \;\ge\; 1-\delta,
  \label{eq:guarantee}
\end{equation}
where $\mathrm{risk}(\lambda)$ is the error rate among the questions a stratum answers at threshold $\lambda$, those scoring strictly above $\lambda$, and the probability is over the calibration draw. The stratification $g$ was fixed before any calibration data were seen: $S_1$ (unanimous first round), scored by $z(\keeprate)+z(\rse)$; $S_2$ (contested but converged), scored by the trajectory-averaged agreement; and $S_3$ (never converged), which abstains as a whole. Within each certified stratum we select the threshold by Learn-then-Test \citep{angelopoulos2021ltt}: each candidate coverage level $c$ on a grid places a threshold $\lambda_c$ at the top-$c$ quantile of the calibration scores, and its selective risk is tested against the null $H_0 : \mathrm{risk}(\lambda_c) > \alpha$ with the exact binomial tail
\begin{equation}
  p_c \;=\; \Pr\big[\,\mathrm{Bin}(n_c, \alpha) \le E_c\,\big],
  \label{eq:ltt}
\end{equation}
where $n_c$ is the number of calibration questions scoring strictly above $\lambda_c$ and $E_c$ the number of errors among them; the grid is tested at Bonferroni-corrected level $\nicefrac{\delta}{|\mathcal{C}|}$ (and across certified strata), and $\hat{\lambda}$ is the threshold of the largest coverage whose null is rejected (Algorithm~\ref{alg:ltt}, Appendix~\ref{app:algos}).

\begin{assumption}
\label{ass:iid}
Within each stratum, calibration and deployment questions are drawn i.i.d.\ from the same distribution.
\end{assumption}

\begin{theorem}
\label{thm:ltt}
Under Assumption~\ref{ass:iid}, the threshold $\hat{\lambda}$ returned by Algorithm~\ref{alg:ltt} satisfies the guarantee of Eq.~\eqref{eq:guarantee} for every certified stratum. (Proof in Appendix~\ref{app:cert}.)
\end{theorem}

The guarantee does not depend on the quality of the score; a weak score lowers certified coverage but not the validity of the bound. Because the scores entering certification ($\rse$, $\keeprate$, $\bar{a}$) are training-free, the full calibration split is available for threshold selection, and we report certified coverage and realized risk as means over repeated calibration/test splits rather than a single draw.

\section{Experiments and Results}
\label{sec:experiments}

\subsection{Experimental Settings}
\label{sec:setup}

\textbf{Datasets.}
We evaluate on MedQA \citep{jin2020medqa}, the medicine subset of MMLU-Pro \citep{wang2024mmlupro}, and MedMCQA \citep{pal2022medmcqa}, in the form distributed with the substrate. MedXpertQA \citep{zuo2025medxpertqa} is a hard-frontier reference on which substrate accuracy approaches the guessing floor. Specialty labels on MedQA and MedMCQA support Appendix~\ref{app:specialty}.

\textbf{Substrate and backbones.}
All experiments instrument the entropy-guided variant of the substrate at a pinned commit, with its retrieval stack unchanged (BM25 with MedCPT cross-encoder reranking \citep{jin2023medcpt} over MedCorp \citep{xiong2024medrag}). Final answers are scored by the substrate's own rule (strict majority, ties as errors), and a logging layer records every round verbatim. We evaluate Qwen3-8B \citep{yang2025qwen3} as the primary backbone, Llama-3.1-8B-Instruct \citep{grattafiori2024llama}, and the medical reasoning model HuatuoGPT-o1-8B \citep{chen2024huatuogpt}, all under identical decoding configuration and seed (Appendix~\ref{app:impl}).

\textbf{Baselines.}
We compare against the abstention signals available to a deployed consensus system: terminal agreement, first-round agreement \citep{wang2023selfconsistency}, mean sequence log-probability, verbalized confidence \citep{tian2023justask, xiong2024verbalized}, and P(True) \citep{kadavath2022know}. On multiple choice, semantic entropy \citep{kuhn2023semantic, farquhar2024semantic} reduces to answer-distribution entropy, which we report as its representative. As the certification control, first-round agreement is calibrated with the same stratified machinery, following the single-round consistency-plus-conformal recipe \citep{yadkori2024conformal}.

\textbf{Metrics and certification.}
Discrimination is reported as AUROC, selective prediction as risk--coverage curves, and probe behavior as flip and keep rates (Eq.~\eqref{eq:keep}); flip rates carry Wilson intervals and certification numbers carry standard errors over repeated splits. The combined process score is fit by logistic regression under question-level cross-validation with fold-internal preprocessing, and rationale claim pairs are scored by DeBERTa-large fine-tuned on MNLI \citep{he2021deberta} (Appendix~\ref{app:impl}). Certification follows the pre-registered stratified protocol of \S\ref{sec:certify}, calibrated independently per domain on an even calibration/test split, sweeping $\alpha \in \{0.05, 0.10, 0.15, 0.20\}$.

\subsection{Discrimination on the Full Question Stream}
\label{sec:results}

\begin{figure}[t]
  \centering
  \includegraphics[width=0.89\linewidth]{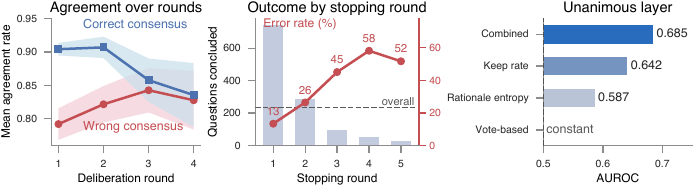}
  \caption{(Left) Correct consensus starts near-unanimous, while wrong consensus starts from disagreement and converges to it slowly (MedQA, mean agreement per round, shaded standard errors). (Middle) Error rate rises with the round at which consensus concludes (bars: questions concluding per round; line: error rate at that round; dashed: overall). (Right) On the unanimous layer every vote-based signal is constant, and only the formation signals discriminate.}
  \label{fig:formation}
\end{figure}

\begin{table}[t]
\centering
\caption{Discrimination (AUROC, higher is better) and selective risk (AURC, lower is better) of abstention signals on the Qwen3-8B substrate. Best value per column in bold. Verbalized confidence and P(True) were elicited on the two primary benchmarks.}
\label{tab:main}
\footnotesize
\setlength{\tabcolsep}{3pt}
\begin{tabular}{lcccccc}
\toprule
& \multicolumn{3}{c}{AUROC $\uparrow$} & \multicolumn{3}{c}{AURC $\downarrow$} \\
\cmidrule(lr){2-4}\cmidrule(lr){5-7}
Signal & MedQA & MMLU-Pro & MedMCQA & MedQA & MMLU-Pro & MedMCQA \\
\midrule
Terminal agreement & \TerminalMedQA & \TerminalMMLUPro & \TerminalMedMCQA & \AurcTerminalMedQA & \AurcTerminalMMLUPro & \AurcTerminalMedMCQA \\
First-round agreement \citep{wang2023selfconsistency} & \ROneMedQA & \ROneMMLUPro & \ROneMedMCQA & \AurcROneMedQA & \AurcROneMMLUPro & \AurcROneMedMCQA \\
Sequence log-probability & \SeqLogprobMedQA & \SeqLogprobMMLUPro & \SeqLogprobMedMCQA & \AurcSeqLogprobMedQA & \AurcSeqLogprobMMLUPro & \AurcSeqLogprobMedMCQA \\
Answer entropy \citep{kuhn2023semantic} & \AnsEntropyMedQA & \AnsEntropyMMLUPro & \AnsEntropyMedMCQA & \AurcAnsEntropyMedQA & \textbf{\AurcAnsEntropyMMLUPro} & \AurcAnsEntropyMedMCQA \\
Verbalized confidence \citep{tian2023justask} & \VerbalizedMedQA & \VerbalizedMMLUPro & -- & \AurcVerbalizedMedQA & \AurcVerbalizedMMLUPro & -- \\
P(True) \citep{kadavath2022know} & \PTrueMedQA & \PTrueMMLUPro & -- & \AurcPTrueMedQA & \AurcPTrueMMLUPro & -- \\
\cmidrule(lr){1-7}
\method{} (process score) & \textbf{\CombinedMedQA} & \textbf{\CombinedMMLUPro} & \textbf{\CombinedMedMCQA} & \textbf{\AurcCombinedMedQA} & \AurcCombinedMMLUPro & \textbf{\AurcCombinedMedMCQA} \\
\bottomrule
\end{tabular}
\end{table}

\begin{table}[t]
\centering
\caption{Process score versus terminal agreement across backbones. \method{} separates correct from incorrect consensus at \CombinedMedQA--\CombinedMedQALlama{} AUROC where terminal agreement is chance, and has the lower selective risk in every cell.}
\label{tab:backbones}
\small
\setlength{\tabcolsep}{3.5pt}
\begin{tabular}{lcccccccc}
\toprule
& \multicolumn{4}{c}{MedQA} & \multicolumn{4}{c}{MMLU-Pro med} \\
\cmidrule(lr){2-5}\cmidrule(lr){6-9}
& \multicolumn{2}{c}{AUROC $\uparrow$} & \multicolumn{2}{c}{AURC $\downarrow$} & \multicolumn{2}{c}{AUROC $\uparrow$} & \multicolumn{2}{c}{AURC $\downarrow$} \\
Backbone & \method{} & Term. & \method{} & Term. & \method{} & Term. & \method{} & Term. \\
\midrule
Qwen3-8B & \textbf{\CombinedMedQA} & \TerminalMedQA & \textbf{\AurcCombinedMedQA} & \AurcTerminalMedQA & \textbf{\CombinedMMLUPro} & \TerminalMMLUPro & \textbf{\AurcCombinedMMLUPro} & \AurcTerminalMMLUPro \\
HuatuoGPT-o1-8B & \textbf{\CombinedMedQAHuatuo} & \TerminalMedQAHuatuo & \textbf{\AurcCombinedMedQAHuatuo} & \AurcTerminalMedQAHuatuo & \textbf{\CombinedMMLUProHuatuo} & \TerminalMMLUProHuatuo & \textbf{\AurcCombinedMMLUProHuatuo} & \AurcTerminalMMLUProHuatuo \\
Llama-3.1-8B & \textbf{\CombinedMedQALlama} & \TerminalMedQALlama & \textbf{\AurcCombinedMedQALlama} & \AurcTerminalMedQALlama & \textbf{\CombinedMMLUProLlama} & \TerminalMMLUProLlama & \textbf{\AurcCombinedMMLUProLlama} & \AurcTerminalMMLUProLlama \\
\bottomrule
\end{tabular}
\end{table}

Figure~\ref{fig:formation} shows that wrong and correct consensus form differently. Correct consensus starts near-unanimous; wrong consensus starts from disagreement, and the error rate rises from \UaErrRateMedQA\% at instant consensus to \FormationRFourErr\% by round four. Table~\ref{tab:main} reports how well each signal separates correct from incorrect final answers. Terminal agreement, the deployed default, is at chance on every benchmark. This follows from the early-stopping rule (\S\ref{sec:process}). The process score, computed from the same runs, raises discrimination to \CombinedMedQA{} on MedQA, \CombinedMMLUPro{} on MMLU-Pro medicine, and \CombinedMedMCQA{} on MedMCQA, and yields the lowest selective risk on MedQA and MedMCQA (risk-coverage curves in Appendix~\ref{app:full}). The remaining single-round signals fall in a narrow band, which the process score exceeds on every benchmark. This aggregate comparison averages over the unanimous layer, which \S\ref{sec:blindspot} examines separately.

The other backbones behave alike (Table~\ref{tab:backbones}): the largest gap is on Llama-3.1-8B, where the process score reaches \CombinedMedQALlama{} AUROC against \TerminalMedQALlama{} for terminal agreement. HuatuoGPT-o1-8B preserves the ordering, and the process score has the lower AURC in all six cells, less than half of terminal agreement's on Llama/MedQA.

\subsection{The Unanimous Layer}
\label{sec:blindspot}

\begin{figure}[t]
  \centering
  \includegraphics[width=0.89\linewidth]{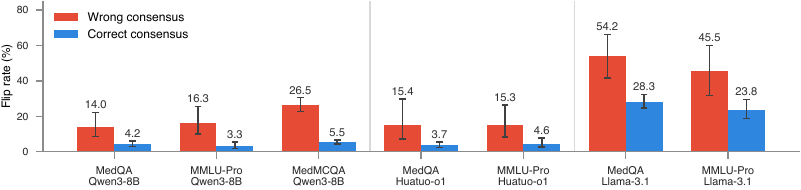}
  \caption{Flip rates of wrong and correct consensus under counter-evidence. A consensus counts as flipped when the majority vote of the $N{=}8$ re-answers no longer lands on the original answer. Flip rates per backbone and benchmark with Wilson 95\% confidence intervals. The asymmetry holds in every cell, and the per-question keep rate discriminates on all seven (AUROC \KeepMMLUPro--\KeepMedMCQA).}
  \label{fig:flip}
\end{figure}

Questions on which all eight candidates agree in the first round are the ones where a wrong answer is hardest to catch. This layer covers \UaRateMedQA\%, \UaRateMMLUPro\%, and \UaRateMedMCQA\% of MedQA, MMLU-Pro medicine, and MedMCQA, where \UaErrRateMedQA\%, \UaErrRateMMLUPro\%, and \UaErrRateMedMCQA\% of unanimous answers are wrong. The error rate grows with difficulty; on expert-frontier MedXpertQA, where \UaRateMedXpert\% of questions still reach first-round unanimity, \UaErrRateMedXpert\% of those unanimous votes are wrong.

Rationale semantic entropy, computed from the existing rationales, separates correct from incorrect unanimity at \FThreeMedQA{} AUROC on MedQA and \FThreeMMLUPro{} on MMLU-Pro medicine, on a subpopulation where the answers themselves are identical.

The counter-evidence probe separates the two more sharply. Figure~\ref{fig:flip} shows each unanimous consensus re-asked with retrieved counter-evidence in context: wrong consensus flips at \FlipWrongMedQA--\FlipWrongMedMCQA\% across the three benchmarks, while correct consensus stays at \FlipRightMMLUPro--\FlipRightMedMCQA\%. The per-question keep rate scores this asymmetry (\KeepMedQA, \KeepMMLUPro, and \KeepMedMCQA{} AUROC on the three benchmarks), and combining it with rationale entropy lifts the unanimous layer to \ComboUaMedQA{} and \ComboUaMMLUPro{} on the two primary benchmarks. Sequence log-probability, at chance or inverted on the full stream (Table~\ref{tab:main}), recovers to \SlpUaMedQA{} inside the layer, while the self-report baselines stay behind (P(True) \PTrueUaMedQA, verbalized confidence \VerbUaMedQA). Adding it to the keep-and-entropy pair lifts the layer to \ComboThreeUaMedQA{} AUROC, so the likelihood and process signals are complementary.

Control arms on MedQA attribute the flips to the retrieved content (Appendix~\ref{app:probectrl}). The devil's-advocate instruction alone flips none of the 746 consensus votes, irrelevant documents flip \FlipWrongCtrlRanddocs\% of wrong consensus, and the counter-evidence itself under a neutral instruction retains nearly the full separation (\FlipWrongCtrlNeutral\% versus \FlipRightCtrlNeutral\%).

The asymmetry recurs on the other backbones. On HuatuoGPT-o1-8B, whose MedQA unanimous layer errs at only \UaErrRateMedQAHuatuo\%, wrong consensus still flips at \FlipWrongMedQAHuatuo\% against \FlipRightMedQAHuatuo\% for correct. Llama-3.1-8B revises far more readily and flips even correct consensus at up to \FlipRightMedQALlama\%, but wrong consensus flips at \FlipWrongMedQALlama\%. Across all seven cells of Figure~\ref{fig:flip} the keep-rate AUROC stays within \KeepMMLUPro--\KeepMedMCQA.

\subsection{Certification Results}
\label{sec:cert}

We now calibrate these scores to the guarantee of Eq.~\eqref{eq:guarantee}. Without stratification, naive quantile calibration on the full question stream misses its target (realized test risk \NaiveViolation\% at $\alpha{=}0.10$ on MedQA), and Learn-then-Test applied globally is valid but returns no certified threshold at any $\alpha$ we sweep. Both fail for the heterogeneity described in \S\ref{sec:certify}: no single threshold on the pooled stream is both valid and non-trivial.

With per-stratum certification, on the MedQA unanimous layer the combined keep-and-entropy score certifies \CertCovMean{}$\pm$\CertCovSE{} of questions at $\alpha{=}\CertAlpha$ over \CertRepeats{} splits, with a realized risk on answered questions of \CertRisk\%, well below the bound. Every vote-based signal of Table~\ref{tab:main} is constant on this stratum, so none of them can certify non-zero coverage here. Multiplied by the layer's share of the benchmark, this certified region amounts to about \CertGlobalShare\% of all MedQA questions. Enlarging the calibration side with in-domain unanimous questions ($n{=}\FuelN$) lifts certified coverage at $\alpha{=}0.15$ to \FuelCertCovFifteen{}$\pm$\FuelCertCovFifteenSE{} and certifies $\alpha{=}0.10$ for the first time (\FuelCertCovTen{}$\pm$\FuelCertCovTenSE, realized risk \FuelCertRiskTen\%), with the test half untouched. On the other backbones, without augmentation, HuatuoGPT-o1-8B certifies \CertCovTenHuatuo{} of its unanimous layer at $\alpha{=}0.10$ and \CertCovFifteenHuatuo{} at $\alpha{=}0.15$, and Llama-3.1-8B \CertCovFifteenLlama{} at $\alpha{=}0.15$ (Figure~\ref{fig:cert}, Appendix~\ref{app:cert}).

Calibration is per domain: pooling the calibration sets of MedQA and MMLU-Pro medicine yields an empty certified region at any level, because the two unanimous layers differ in base error rate (\S\ref{sec:blindspot}). Per-domain calibration is also what Assumption~\ref{ass:iid} requires.

\section{Analysis and Discussion}
\label{sec:analysis}

\subsection{Ablation Studies}
\label{sec:ablation}

Only the rationale family is non-redundant: dropping it lowers MedQA AUROC from \CombinedMedQA{} to \AblDropFThree. The trajectory families are redundant with one another; each alone supports an AUROC close to the full score, and each tracks difficulty proxies (retrieval volume, stopping round) at $|\rho| \ge \RhoTrajMin$. Rationale entropy is the exception on both counts ($|\rho| \le \RhoRseMax$): it is the only family whose removal lowers AUROC and the only one that does not track difficulty (full drop-one matrices in Appendix~\ref{app:ablation}).

Inside the unanimous layer, the two evidence signals are complementary: their combination outperforms either alone (\S\ref{sec:blindspot}). Replacing hard entailment clustering with the continuous score of Eq.~\eqref{eq:rse} raises the unanimous-layer AUROC from \HardFThreeMedQA{} to \FThreeMedQA{} (MedQA) and \HardFThreeMMLUPro{} to \FThreeMMLUPro{} (MMLU-Pro medicine).

First-round agreement isolates the effect of aggregating votes without any process information; inside the unanimous layer, which carries \UaErrNMedQA{} wrong answers on MedQA alone, it is constant, and the separation reported in \S\ref{sec:blindspot} comes entirely from the process signals.

\subsection{Cost and Sensitivity}
\label{sec:cost}

The probe fires only on unanimous first-round votes, at a median of \ProbeMedianSecMedQA{}\,s and \ProbeMedianTok{} completion tokens on MedQA, against \GenMedianSecMedQA{}\,s for generation itself (full accounting in Appendix~\ref{app:cost}); all passive features come from logs the system already writes. Excluding the probe-pilot questions from evaluation shifts the combined AUROC by \SmokeDelta, and gradient-boosted trees in place of logistic regression change no conclusion (Appendix~\ref{app:ablation}).

\subsection{Discussion and Limitations}
\label{sec:discussion}

A hospital deploying a consensus QA system would calibrate \method{} on its own case mix and refer every question outside the certified region to a clinician. Each referred question carries its challenge documents and the record of dissent, so the clinician sees the evidence that led the system to withhold an answer. The signals are defined for any system that deliberates in rounds and require nothing beyond its execution log. Tighter certificates require more calibration data, and in-domain unanimous questions accumulate as the system runs (\S\ref{sec:cert}). The guarantee applies in any stratum whose base error rate is at or below the target $\alpha$; a stratum above it receives no certified region.

The evaluation is limited to multiple-choice benchmarks, so whether the flip asymmetry and the rationale signals carry over to free-text clinical questions is untested. The three backbones are 8B-scale models on a single published substrate; the signal definitions transfer to other deliberating systems, but we have not measured them elsewhere. The certificates hold under Assumption~\ref{ass:iid}, so a deployment must calibrate on its own case mix and recalibrate when that mix drifts. And we have not evaluated whether clinicians find the returned dissent records useful; a prospective evaluation with clinicians is the next step.

\section{Conclusion}
\label{sec:conclusion}

We presented \method{}, a certified abstention framework for multi-round consensus medical QA. Its abstention decision is computed from the record of how the agreement was reached: process features from the execution log, rationale entropy and a counter-evidence probe for unanimous votes, and a stratified Learn-then-Test calibration of the resulting scores at a user-specified selective-risk level. Across three backbones and three benchmarks, terminal agreement is at chance, the process signals separate correct from incorrect consensus on the full stream and inside the unanimous layer, and stratified calibration certifies answering on part of that layer at the target risk.

\bibliographystyle{plainnat}
\bibliography{references}

\newpage
\appendix

\section{Prompts}\label{app:prompts}
This appendix reproduces every prompt in the pipeline verbatim. The first group belongs to the substrate and is unchanged from the pinned commit; we list it for self-containment, and because two of its design choices matter for interpreting our results. The second group is ours: the two probe templates and the two self-report elicitation prompts. Template variables appear in double braces.

\paragraph{Substrate prompts.}
Candidate answering uses a plain system instruction with a tagged-answer format; the same system prompt is reused by later rounds and by our probe re-answering, so no probe effect can be attributed to a change of role or format.

\begin{promptbox}{Candidate answering (system, all rounds)}
\small\ttfamily
You are a medical assistant, please answer my medical questions. Give your final choice (capital option) closed in tag `<answer>X</answer>` after your analysis.\\
Your response should be as detailed as possible, but please do not use any subheadings.
\end{promptbox}

\begin{promptbox}{Candidate answering (user, round 1)}
\small\ttfamily
Below is a multiple-choice question.\\
\#\#\# Question\\
\{\{question\}\}\\[2pt]
\#\#\# Options\\
\{\{options\}\}\\[2pt]
Please analyze the question and give your answer. Give your analysis and final choice.
\end{promptbox}

Two features of the substrate's later-round prompt bear directly on \S\ref{sec:results}. First, candidates are shown the previous answers sorted by token-entropy confidence, together with an explicit caveat that the ranking may be wrong. Second, the prompt closes by instructing the model to answer \emph{towards lower entropy}: convergence is not only an emergent behavior of the panel but an instructed objective. A system whose members are told to converge will converge on wrong answers as readily as on right ones, which is one more reason the terminal agreement it produces carries no information about correctness.

\begin{promptbox}{Evidence-conditioned re-answering (user, rounds $\ge$ 2)}
\small\ttfamily
Below is a multiple-choice question.\\
\#\#\# Question\\
\{\{question\}\}\\[2pt]
\#\#\# Options\\
\{\{options\}\}\\[2pt]
\#\#\# Documents\\
\{\{documents\}\}\\[2pt]
---\\
I will provide several assistant's previous answers, which may be incorrect.\\
The previous answers are sorted by their confidence (low entropy means high confidence), but the entropy metric cannot accurately reflect whether the answer is correct or not.\\
Please analyze the previous answers and then re-answer.\\[2pt]
\{\{answers\}\}\\[2pt]
---\\
The above are the question along with the assistant's previous answers, and some relevant documents. Please analyze and then re-answer. Please give your response towards **lower entropy**. Give your analysis and final answer.
\end{promptbox}

When the panel disagrees, the substrate generates retrieval queries from the disagreement itself. The query prompt asks for the \emph{dispute points} among the answers, which is precisely the structure our counter-query prompt inverts.

\begin{promptbox}{Conflict-guided query generation (system)}
\small\ttfamily
\#\#\# Role \& Goal:\\
You are a medical expert. I will provide multiple different answers to the same question, which may be incorrect. Your task is to identify contradictions, ambiguities, and core dispute points among the answers, and extract key concept queries for further retrieval and verification.\\[2pt]
\#\#\# Processing Steps:\\
1. Analyze different answers and summarize the differences between them.\\
2. Extract keywords from these differences for retrieval.\\
3. Generate 1-4 precise queries for further search using **BM25** retriever.\\[2pt]
\#\#\# Output format:\\
ONLY give the queries:\\
{[}Query 1{]} xxx\\
{[}Query 2{]} xxx\\
(more queries)
\end{promptbox}

\paragraph{Probe prompts.}
The counter-query prompt keeps the substrate's query-generation structure (role header, numbered steps implied by the format, BM25-targeted output) and changes only the direction of search: instead of resolving a disagreement, it retrieves evidence against an agreement. Keeping the format aligned means the probe exercises the same retrieval machinery under the same interface, so flip behavior reflects the consensus, not a novel pipeline.

\begin{promptbox}{Counter-evidence query generation (system; thinking mode, temperature 0)}
\small\ttfamily
\#\#\# Role \& Goal:\\
You are a skeptical medical expert. All assistants unanimously chose the same answer to a question. Your task is to actively search for evidence that could CHALLENGE or CONTRADICT that consensus: evidence supporting plausible alternative options, known exceptions, contraindications, or updated guidelines.\\[2pt]
\#\#\# Output format:\\
ONLY give the queries:\\
{[}Query 1{]} xxx\\
{[}Query 2{]} xxx\\
(more queries, 2-4 total, for a **BM25** retriever)
\end{promptbox}

\begin{promptbox}{Counter-evidence query generation (user)}
\small\ttfamily
Below is a multiple-choice question.\\
\#\#\# Question\\
\{\{question\}\}\\[2pt]
\#\#\# Options\\
\{\{options\}\}\\[2pt]
---\\
All assistants unanimously answered: \{\{consensus\}\}.\\[2pt]
Generate 2-4 precise BM25 queries to retrieve evidence that could challenge this consensus (e.g., evidence favoring other options or against \{\{consensus\}\}). Give your queries in the given format.
\end{promptbox}

The re-answering prompt needs the most care. A naive challenge instruction produces sycophantic flipping: models abandon any answer once an authority questions it, which would erase the very asymmetry the probe measures. The wording below therefore allows but does not require a change of answer; the key sentence is \emph{do not change your answer unless the evidence genuinely warrants it}. Under this instruction correct consensus holds at \FlipRightMMLUPro{}--\FlipRightMedMCQA\% while wrong consensus flips at \FlipWrongMedQA{}--\FlipWrongMedMCQA\% (Figure~\ref{fig:flip}), and the control arms of Appendix~\ref{app:probectrl} confirm the division of labor directly: the instruction alone flips nothing, and the counter-evidence carries the asymmetry.

\begin{promptbox}{Devil's-advocate re-answering (user; system prompt identical to candidate answering)}
\small\ttfamily
Below is a multiple-choice question.\\
\#\#\# Question\\
\{\{question\}\}\\[2pt]
\#\#\# Options\\
\{\{options\}\}\\[2pt]
\#\#\# Challenge Documents\\
A skeptical reviewer retrieved the following documents that may challenge the previous consensus:\\
\{\{documents\}\}\\[2pt]
---\\
Previously, all assistants unanimously answered \{\{consensus\}\}. A devil's advocate now asks: in light of the challenge documents above, critically re-examine whether \{\{consensus\}\} is really correct, or whether another option is better supported. Weigh the challenge evidence seriously, but do not change your answer unless the evidence genuinely warrants it. Give your analysis and final choice.
\end{promptbox}

\paragraph{Self-report elicitation prompts.}
Both baselines are elicited post hoc on the system's final answer, one call per question at temperature 0. Verbalized confidence follows the 0--100 elicitation of \citet{tian2023justask, xiong2024verbalized}; P(True) follows \citet{kadavath2022know}, with the score read from the first generated token's top-20 log-probability mass over \emph{True}/\emph{False} variants rather than from the sampled text.

\begin{promptbox}{Verbalized confidence (system + user)}
\small\ttfamily
You are a careful medical assistant. Rate how confident you are that a given answer to a multiple-choice question is correct. Reply with a single integer from 0 to 100 inside <conf></conf> tags, then one short sentence of justification.\\[3pt]
\#\#\# Question\\
\{\{question\}\}\\[2pt]
\#\#\# Options\\
\{\{options\}\}\\[2pt]
\#\#\# Proposed answer: \{\{answer\}\}\\[2pt]
How confident are you (0-100) that this answer is correct? Reply with <conf>N</conf> first.
\end{promptbox}

\begin{promptbox}{P(True) (system + user)}
\small\ttfamily
You are a careful medical assistant. Judge whether a given answer to a multiple-choice question is correct. Reply with exactly one word: True or False.\\[3pt]
\#\#\# Question\\
\{\{question\}\}\\[2pt]
\#\#\# Options\\
\{\{options\}\}\\[2pt]
\#\#\# Proposed answer: \{\{answer\}\}\\[2pt]
Is the proposed answer correct? Reply with exactly one word: True or False.
\end{promptbox}

\section{Feature Definitions}\label{app:features}
The process score is fit over five feature families, each built from a small set of base quantities and their standard temporal aggregations (value at the first and last round, mean, slope, and maximum over rounds). Table~\ref{tab:families} summarizes what each family measures; the representative statistics introduced in \S\ref{sec:process} are the \emph{trajectory-averaged agreement}
\begin{equation}
  \bar{a} \;=\; \tfrac{1}{\Tstop}\,\textstyle\sum_{t=1}^{\Tstop} \st{\mt},
  \label{eq:auc}
\end{equation}
the \emph{majority-flip count}
\begin{equation}
  n_{\mathrm{flip}} \;=\; \bigl|\{\, t < \Tstop : \mt[t+1] \neq \mt \,\}\bigr|,
  \label{eq:flip}
\end{equation}
\emph{minority persistence}
\begin{equation}
  \pi \;=\; \max_{a \neq m^\ast}\; \tfrac{1}{\Tstop} \, \bigl|\{\, t : \st{a} \ge \nicefrac{1}{N} \,\}\bigr|,
  \label{eq:minority}
\end{equation}
where $m^\ast$ is the final answer and the count is the number of rounds in which option $a$ retains at least one supporter, and the \emph{document-repeat rate}
\begin{equation}
  \rho_t \;=\; \bigl|\,D_t \cap \textstyle\bigcup_{t' < t} D_{t'}\bigr| \,\big/\, |D_t|,
  \label{eq:docrepeat}
\end{equation}
with first-round agreement $a_1$ as in Eq.~\eqref{eq:r1} and the rationale representative as in Eq.~\eqref{eq:rse}.
The answer-distribution family is built on the per-round \emph{answer entropy}
\begin{equation}
  H_t \;=\; -\textstyle\sum_{a} \st{a} \ln \st{a},
  \label{eq:ansent}
\end{equation}
aggregated as its first- and last-round values, its slope, and the fraction of rounds in which it does not increase, together with the Jensen--Shannon divergence between the answer distributions of adjacent rounds.

\begin{table}[h]
\centering
\caption{Feature families of the process score.}
\label{tab:families}
\small
\setlength{\tabcolsep}{4.5pt}
\begin{tabular}{llc}
\toprule
Family & Measures & Representative \\
\midrule
Agreement dynamics & how fast and how stably the vote converges & Eqs.~\eqref{eq:r1}, \eqref{eq:auc}, \eqref{eq:flip} \\
Answer-distribution dynamics & entropy and drift of the candidate answer distribution & Eq.~\eqref{eq:ansent} \\
Minority dynamics & whether dissent is persistent or transient & Eq.~\eqref{eq:minority} \\
Rationale signals & mutual entailment among the reasons behind the votes & Eq.~\eqref{eq:rse} \\
Retrieval dynamics & whether deliberation still finds new evidence & Eq.~\eqref{eq:docrepeat} \\
\bottomrule
\end{tabular}
\end{table}

Every definition, the probe protocol, and the stratification were frozen in a versioned registry before the corresponding data were collected; the machine-readable registry with exact formulas for all features ships with the code.

\section{Claim Extraction Rule}\label{app:claims}
Rationale semantic entropy scores mutual entailment among the \emph{concluding claims} of the panel's rationales, not among the rationales themselves. This appendix documents why that restriction is necessary and gives the exact extraction rule.

Scoring full rationales fails in two ways. Rationales run to hundreds of tokens, well past the effective input of the entailment model, so truncation decides what gets compared. More fundamentally, two rationales about the same clinical question share most of their surface vocabulary whether or not they agree: both walk the same options, cite the same anatomy, and eliminate the same distractors. Entailment scores over full rationales therefore saturate on topical similarity, and the logical disagreement, which lives almost entirely in the final verdict sentences, is drowned out. Restricting the comparison to concluding claims is what makes the NLI calls well-posed.

Prior work that decomposes generations into checkable units typically prompts a strong LLM to do the splitting \citep{mohri2024factuality}. We use a deterministic rule instead, for three reasons. A fixed rule can be frozen before any rationale data are inspected, which an LLM extractor cannot guarantee; it adds zero inference cost to a feature set whose selling point is being free; and it avoids a circularity, since asking a model to summarize its own reasoning gives it a second chance to smooth over exactly the incoherence the feature is trying to measure.

The rule itself has four steps and was fixed before any rationale data were inspected. The answer tag is stripped and the remaining text is split into sentences. Within the last five sentences, every sentence that contains a conclusion cue is kept, up to three of them and in their original order; the cue list has nine entries: \emph{answer}, \emph{option}, \emph{choice}, \emph{correct}, \emph{therefore}, \emph{thus}, \emph{hence}, \emph{most likely}, and \emph{best}. When no sentence carries a cue, the final sentence serves as the claim. The implementation and its unit tests ship with the code.

\begin{promptbox}{Worked example (MedQA, dimeric-immunoglobulin question, candidate 2)}
\small
\textbf{Rationale (final portion of 1{,}367 characters):} ``\ldots Option C, `Inhibiting bacterial adherance and colonization of mucous membranes,' is the most accurate description of the primary function of dimeric IgA. \texttt{<answer>C</answer>}''\\[3pt]
\textbf{Extracted claim:} ``Option C, `Inhibiting bacterial adherance and colonization of mucous membranes,' is the most accurate description of the primary function of dimeric IgA.''
\end{promptbox}

The rule is deliberately permissive: when several tail sentences carry cues, all of them survive, including option-elimination lines. This costs precision on individual extractions but keeps the rule simple, auditable, and free of tuned thresholds, and the resulting score separates knowledge-backed from inherited unanimity at \FThreeMedQA{} and \FThreeMMLUPro{} AUROC on the two primary benchmarks (\S\ref{sec:blindspot}).

\section{Algorithms}\label{app:algos}
Algorithm~\ref{alg:probe} gives the probe of \S\ref{sec:evidence}; each line is one call into machinery the substrate already runs, with the prompts of Appendix~\ref{app:prompts}. Algorithm~\ref{alg:ltt} gives the calibration of \S\ref{sec:certify}; following \citet{quach2024clm}, the statistical content stays in Eqs.~\eqref{eq:guarantee} and~\eqref{eq:ltt}, and the box only fixes the order of operations.

\begin{algorithm}[h]
\caption{Counter-evidence probe}\label{alg:probe}
\begin{algorithmic}[1]
\REQUIRE question $x$ with unanimous first-round consensus $m^\ast$ ($a_1 = 1$); panel size $N$; retriever $\mathcal{R}$ of the substrate; query budget $k \le 4$
\ENSURE keep rate $\keeprate \in \{0, \nicefrac{1}{N}, \dots, 1\}$
\STATE $q_1, \dots, q_j \gets \textsc{CounterQueries}(x, m^\ast)$, $j \le k$ \hfill $\triangleright$ temperature 0, thinking mode
\STATE $D_{\mathrm{ch}} \gets \mathcal{R}(q_1, \dots, q_j)$ \hfill $\triangleright$ same BM25 + rerank stack, $\le 8$ documents
\FOR{$i = 1$ \TO $N$}
\STATE $\tilde{a}_i \gets \textsc{ReAnswer}(x, D_{\mathrm{ch}}, m^\ast)$ \hfill $\triangleright$ devil's-advocate instruction; revision licensed, not demanded
\ENDFOR
\STATE $\keeprate \gets \bigl|\{\, i : \tilde{a}_i = m^\ast \,\}\bigr| \,/\, N$
\RETURN $\keeprate$ \hfill $\triangleright$ low $\keeprate$: the consensus collapses under challenge
\end{algorithmic}
\end{algorithm}

\begin{algorithm}[h]
\caption{Stratified Learn-then-Test calibration}\label{alg:ltt}
\begin{algorithmic}[1]
\REQUIRE calibration set $\mathcal{D}_{\mathrm{cal}}$; stratification $g$ with certified strata $\{S_1, S_2\}$; per-stratum scores $s_S$; risk level $\alpha$; error budget $\delta$; coverage grid $\mathcal{C}$
\ENSURE thresholds $\hat{\lambda}_S$, or \textsc{abstain} for strata certifying nothing
\FOR{each certified stratum $S$}
\STATE $\mathcal{D}_S \gets \{\, (x, y) \in \mathcal{D}_{\mathrm{cal}} : g(x) = S \,\}$
\FOR{$c \in \mathcal{C}$ in descending order}
\STATE $\lambda_c \gets$ score at the top-$c$ quantile of $\mathcal{D}_S$
\STATE $n_c \gets \bigl|\{\, x \in \mathcal{D}_S : s_S(x) > \lambda_c \,\}\bigr|$; \quad $E_c \gets$ errors among them
\STATE $p_c \gets \Pr\bigl[\mathrm{Bin}(n_c, \alpha) \le E_c\bigr]$ \hfill $\triangleright$ Eq.~\eqref{eq:ltt}
\IF{$p_c \le \delta \,/\, (|\mathcal{C}| \cdot |\{S_1, S_2\}|)$}
\STATE $\hat{\lambda}_S \gets \lambda_c$; \textbf{break} \hfill $\triangleright$ largest certified coverage for $S$
\ENDIF
\ENDFOR
\STATE \textbf{if} no rejection \textbf{then} $\hat{\lambda}_S \gets$ \textsc{abstain} \hfill $\triangleright$ stratum uncertified at level $\alpha$
\ENDFOR
\RETURN $\{\hat{\lambda}_S\}$ \hfill $\triangleright$ answer $x \in S$ iff $s_S(x) > \hat{\lambda}_S$; $S_3$ abstains as a whole
\end{algorithmic}
\end{algorithm}

\section{Certification Protocol and Proof}\label{app:cert}

\paragraph{Protocol.}
The coverage grid runs from $0.05$ to $1.00$ in steps of $0.05$, giving twenty candidate thresholds per stratum. The family-wise error budget is $\delta = 0.05$, Bonferroni-corrected across the twenty grid points and the two certified strata, and we sweep $\alpha \in \{0.05, 0.10, 0.15, 0.20\}$, each level constituting its own guarantee. Calibration and test halves are split evenly at random, and every certification number in the paper is the mean over \CertRepeats{} such splits with fixed seeds: a single split yields a valid certificate, but its reported coverage inherits the variance of one draw, and averaging over repeated splits reports the procedure rather than one particular partition. The per-split violation fraction is the quantity the $1-\delta$ guarantee bounds, and it is \CertViolations{} of \CertRepeats{} for every Learn-then-Test arm reported: no split's realized test risk exceeds its target $\alpha$. Calibration is per domain throughout; \S\ref{sec:cert} reports what happens when domains are pooled.

\paragraph{Why a grid rather than fixed-sequence testing.}
Selective risk is not monotone in the threshold, so the fixed-sequence variant of Learn-then-Test, which walks thresholds from the most selective downward and stops at the first failure, is brittle here: one wrong answer among the few highest-scoring calibration questions terminates the walk at zero certified coverage even when broader coverage levels are certifiable. Testing the whole grid under a Bonferroni correction is robust to the location of individual errors and costs only a constant factor in the effective test level.

\paragraph{Proof of Theorem~\ref{thm:ltt}.}
Fix a certified stratum with calibration pairs $(s_i, y_i)_{i \le n}$ of score and correctness, and fix a grid point $c$. The threshold $\lambda_c$ is an order statistic of the calibration scores, so the argument conditions on it. Let $A = \{\, i : s_i > \lambda_c \,\}$ be the calibration questions above the threshold, so that $n_c = |A|$. The event that $\lambda_c$ equals a value $t$ and that the set above it equals $A$ constrains the pairs in $A$ only through $s_i > t$, and places all its remaining conditions on the pairs outside $A$. Under Assumption~\ref{ass:iid} the pairs are independent, so conditional on $(\lambda_c, A)$ the pairs in $A$ are i.i.d.\ draws from the distribution of a question given $s > \lambda_c$. This is the population that the rule answers at threshold $\lambda_c$, and its error rate is $\mathrm{risk}(\lambda_c)$. Conditional on $(\lambda_c, A)$, the error count therefore satisfies $E_c \sim \mathrm{Bin}(n_c, \mathrm{risk}(\lambda_c))$. Whenever $\mathrm{risk}(\lambda_c) > \alpha$, this count stochastically dominates $\mathrm{Bin}(n_c, \alpha)$, and the tail probability of Eq.~\eqref{eq:ltt} obeys $\Pr[\, p_c \le u \mid \lambda_c, A \,] \le u$ for every $u \in [0,1]$. Taking the expectation over $(\lambda_c, A)$ gives $\Pr[\, \mathrm{risk}(\lambda_c) > \alpha,\; p_c \le u \,] \le u$, which makes $p_c$ a valid $p$-value for $H_0^{(c)} : \mathrm{risk}(\lambda_c) > \alpha$ at the data-dependent threshold. Algorithm~\ref{alg:ltt} rejects only when $p_c \le \delta / (|\mathcal{C}| \cdot 2)$, so by the union bound over all grid points and both certified strata, the probability that any true null is rejected is at most $\delta$. On the complementary event, of probability at least $1 - \delta$, every rejected null is false; in particular the selected $\hat{\lambda}$, which corresponds to a rejected null, satisfies $\mathrm{risk}(\hat{\lambda}) \le \alpha$. The union-bound step is the Learn-then-Test argument of \citet{angelopoulos2021ltt} with the exact binomial tail. The conditioning step extends it to thresholds placed at calibration quantiles, and it is why the test counts the questions strictly above the threshold: the question that defines the threshold is not a draw from the answered population. \hfill $\square$

\paragraph{Scope of the guarantee.}
The guarantee is marginal over the calibration draw and holds per stratum and per domain; it is distribution-free in the sense that no assumption is placed on the score. Training-free scores are what make the full calibration split available for threshold selection: no data are spent fitting a scorer.

\begin{figure}[h]
  \centering
  \includegraphics[width=\linewidth]{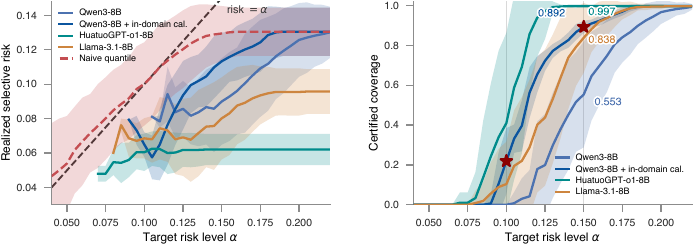}
  \caption{Realized risk and certified coverage of the stratified rule. (Left) Realized selective risk against the target level: every Learn-then-Test arm stays below the risk${}={}\alpha$ diagonal at every level it certifies, while naive quantile calibration crosses it. (Right) Certified coverage per backbone on the MedQA unanimous layer; printed values are the $\alpha{=}0.15$ coverages, and stars mark the two operating points of \S\ref{sec:cert}. Lines are means over \CertRepeats{} calibration/test splits; bands are $\pm$1 standard deviation.}
  \label{fig:cert}
\end{figure}

\section{Full Per-Dataset Results}\label{app:full}
Table~\ref{tab:full} extends the main results to every signal, benchmark, and backbone, reporting both discrimination and selective risk. Unanimous-layer metrics, computed on a different subpopulation, are in \S\ref{sec:blindspot} and Figure~\ref{fig:flip} rather than here.

The full grid confirms the structure of \S\ref{sec:results} without exception. Terminal agreement is at chance in all seven cells, on the strongest backbone as on the weakest. Sequence log-probability is the only signal that gets \emph{worse} than chance, and its selective risk degrades sharply on Llama-3.1-8B, where long free-text rationales make raw likelihood an actively misleading confidence proxy. Answer-distribution entropy tracks first-round agreement closely everywhere, as both are functionals of the same round-one vote distribution. The process score has the best discrimination in six of the seven cells and the best selective risk in five, and its margin over the strongest snapshot signal is widest where the backbone is weakest.

\begin{figure}[h]
  \centering
  \includegraphics[width=\linewidth]{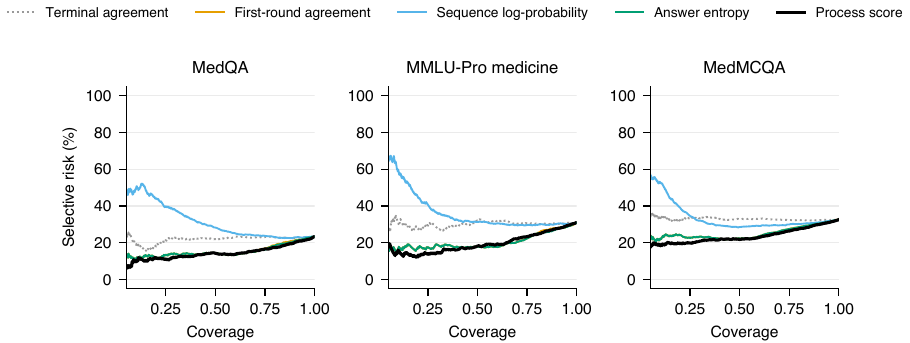}
  \caption{Risk-coverage curves of the always-available signals on the Qwen3-8B substrate; lower is better, and the AURC values of Table~\ref{tab:main} are the areas under these curves. The process score uses the same five-fold out-of-fold protocol as the tables; coverage below 5\% is omitted.}
  \label{fig:rc}
\end{figure}

Figure~\ref{fig:rc} plots the risk-coverage curves behind the AURC column on the Qwen3-8B substrate. The process score traces the lowest risk over nearly the whole coverage range on all three benchmarks, and the ordering of the other signals is stable: terminal agreement is flat, sequence log-probability is inverted at low coverage, and the two round-one functionals move together.

\begin{table}[h]
\centering
\caption{Discrimination (AUROC) and selective risk (AURC) of all signals across benchmarks and backbones. Dashes mark cells outside the evaluation design: the cross-backbone replication covers the two primary benchmarks.}
\label{tab:full}
\footnotesize
\setlength{\tabcolsep}{3.5pt}
\begin{tabular}{lcccccc}
\toprule
& \multicolumn{3}{c}{AUROC $\uparrow$} & \multicolumn{3}{c}{AURC $\downarrow$} \\
\cmidrule(lr){2-4}\cmidrule(lr){5-7}
Signal & MedQA & MMLU-Pro & MedMCQA & MedQA & MMLU-Pro & MedMCQA \\
\midrule
\multicolumn{7}{l}{\emph{Qwen3-8B}} \\
Terminal agreement & 0.512 & 0.504 & 0.507 & 0.221 & 0.300 & 0.330 \\
First-round agreement \citep{wang2023selfconsistency} & 0.677 & 0.698 & 0.640 & 0.155 & 0.200 & 0.250 \\
Sequence log-probability & 0.400 & 0.474 & 0.534 & 0.320 & 0.376 & 0.347 \\
Answer entropy \citep{kuhn2023semantic} & 0.681 & 0.703 & 0.641 & 0.154 & 0.199 & 0.250 \\
Verbalized confidence \citep{tian2023justask} & 0.585 & 0.651 & -- & 0.179 & 0.207 & -- \\
P(True) \citep{kadavath2022know} & 0.602 & 0.644 & -- & 0.160 & 0.205 & -- \\
\cmidrule(lr){1-7}
\method{} (process score) & 0.696 & 0.704 & 0.659 & 0.145 & 0.205 & 0.234 \\
\midrule
\multicolumn{7}{l}{\emph{HuatuoGPT-o1-8B}} \\
Terminal agreement & 0.519 & 0.526 & -- & 0.193 & 0.353 & -- \\
First-round agreement \citep{wang2023selfconsistency} & 0.766 & 0.701 & -- & 0.093 & 0.232 & -- \\
Sequence log-probability & 0.350 & 0.422 & -- & 0.265 & 0.399 & -- \\
Answer entropy \citep{kuhn2023semantic} & 0.768 & 0.707 & -- & 0.093 & 0.230 & -- \\
\cmidrule(lr){1-7}
\method{} (process score) & 0.775 & 0.701 & -- & 0.096 & 0.225 & -- \\
\midrule
\multicolumn{7}{l}{\emph{Llama-3.1-8B}} \\
Terminal agreement & 0.548 & 0.551 & -- & 0.289 & 0.406 & -- \\
First-round agreement \citep{wang2023selfconsistency} & 0.781 & 0.736 & -- & 0.159 & 0.254 & -- \\
Sequence log-probability & 0.205 & 0.217 & -- & 0.547 & 0.640 & -- \\
Answer entropy \citep{kuhn2023semantic} & 0.790 & 0.744 & -- & 0.156 & 0.252 & -- \\
\cmidrule(lr){1-7}
\method{} (process score) & 0.825 & 0.782 & -- & 0.142 & 0.246 & -- \\
\bottomrule
\end{tabular}

\end{table}

\section{Probe Control Arms}\label{app:probectrl}
Table~\ref{tab:probectrl} isolates what drives the flip asymmetry of \S\ref{sec:blindspot}. Each control arm re-asks the same 746 unanimous MedQA questions with the same $N{=}8$ re-answers, varying one factor at a time: the instruction-only arm keeps the devil's-advocate framing and removes all documents; the irrelevant-documents arm keeps the framing and substitutes documents retrieved for other questions' counter-queries; the neutral-instruction arm keeps this question's counter-evidence documents and removes the adversarial framing.

The instruction alone flips no majority at all, and documents without evidential content flip \FlipWrongCtrlRanddocs\% of wrong consensus. The counter-evidence under a neutral instruction retains nearly the full asymmetry (\FlipWrongCtrlNeutral\% versus \FlipRightCtrlNeutral\%), and the paired difference in asymmetry between the probe and each content-free arm excludes zero (bootstrap 95\% intervals $[\CtrlDiffNodocsLo, \CtrlDiffNodocsHi]$ and $[\CtrlDiffRanddocsLo, \CtrlDiffRanddocsHi]$ percentage points). The flip responds to the content of the counter-evidence, and the probe inherits its discrimination from that content.

\begin{table}[h]
\centering
\caption{Flip rates of the counter-evidence probe and its control arms on the MedQA unanimous layer (Qwen3-8B, $N{=}8$ re-answers per question).}
\label{tab:probectrl}
\small
\setlength{\tabcolsep}{5pt}
\begin{tabular}{lllcc}
\toprule
\textbf{Arm} & \textbf{Instruction} & \textbf{Documents} & \textbf{flip(wrong)} & \textbf{flip(right)} \\
\midrule
Counter-evidence probe & devil's advocate & counter-evidence & \FlipWrongMedQA\% & \FlipRightMedQA\% \\
Instruction only & devil's advocate & none & \FlipWrongCtrlNodocs\% & \FlipRightCtrlNodocs\% \\
Irrelevant documents & devil's advocate & unrelated & \FlipWrongCtrlRanddocs\% & \FlipRightCtrlRanddocs\% \\
Neutral instruction & neutral & counter-evidence & \FlipWrongCtrlNeutral\% & \FlipRightCtrlNeutral\% \\
\bottomrule
\end{tabular}
\end{table}

\section{Ablation Matrices}\label{app:ablation}
Table~\ref{tab:ablation} extends the drop-one ablation of \S\ref{sec:ablation} to all three benchmarks, removing one feature family at a time from the full process score, and adds the classifier swap.

\begin{table}[h]
\centering
\caption{Drop-one family ablation and classifier swap (AUROC, Qwen3-8B substrate). Each row removes the feature family marked \xmark{} from the full process score (top row).}
\label{tab:ablation}
\small
\begin{tabular}{cccccccc}
\toprule
\multicolumn{5}{c}{Feature family} & \multicolumn{3}{c}{AUROC} \\
\cmidrule(lr){1-5}\cmidrule(lr){6-8}
Agreement & Distribution & Minority & Rationale & Retrieval & MedQA & MMLU-Pro & MedMCQA \\
\midrule
\cmark & \cmark & \cmark & \cmark & \cmark & \textbf{0.696} & \textbf{0.704} & \textbf{0.659} \\
\xmark & \cmark & \cmark & \cmark & \cmark & 0.703 & 0.705 & 0.660 \\
\cmark & \xmark & \cmark & \cmark & \cmark & 0.703 & 0.714 & 0.657 \\
\cmark & \cmark & \xmark & \cmark & \cmark & 0.697 & 0.705 & 0.658 \\
\cmark & \cmark & \cmark & \xmark & \cmark & 0.682 & 0.674 & 0.645 \\
\cmark & \cmark & \cmark & \cmark & \xmark & 0.700 & 0.697 & 0.658 \\
\midrule
\multicolumn{5}{l}{GBT in place of logistic regression} & 0.644 & 0.643 & 0.619 \\
\bottomrule
\end{tabular}

\end{table}

The pattern of \S\ref{sec:ablation} holds on every benchmark, not only MedQA: removing the rationale family costs discrimination in all three columns, while removing any other family leaves the score unchanged or marginally improved. The trajectory families are informative individually but interchangeable collectively, each a different projection of the same consensus dynamics; the rationales are the one signal source the rest cannot substitute for. The classifier swap points the same direction on all benchmarks as well, so the choice of a default logistic regression is not doing hidden work. The hard-versus-soft aggregation comparison for the rationale score is in \S\ref{sec:ablation}, and the pilot-question sensitivity check in \S\ref{sec:cost}.

\section{Specialty-Level Certification}\label{app:specialty}
Specialty labels are available for 899 MedQA test questions, of which 529 fall in the unanimous layer. Table~\ref{tab:specialty} breaks the layer-level certificate of \S\ref{sec:cert} down by specialty, using the same score, level, and repeated-split protocol; specialties with fewer than 25 unanimous questions are omitted.

\begin{table}[h]
\centering
\caption{Per-specialty breakdown of the certified rule on the MedQA unanimous layer ($\alpha{=}\CertAlpha$, \CertRepeats{} splits). The certificate itself is calibrated at the layer level; this table reports how its coverage and realized risk distribute across specialties.}
\label{tab:specialty}
\small
\begin{tabular}{lcccc}
\toprule
Specialty & $n$ & Unanimous-layer error (\%) & Certified coverage & Realized risk (\%) \\
\midrule
Cardiology & 42 & 14.3 & 0.70 & 9.0 \\
Hematology & 29 & 17.2 & 0.69 & 13.3 \\
Obstetrics and gynecology & 51 & 11.8 & 0.66 & 14.5 \\
Gastroenterology & 57 & 15.8 & 0.60 & 11.1 \\
Pediatrics & 35 & 11.4 & 0.60 & 8.9 \\
Psychiatry & 33 & 12.1 & 0.59 & 7.3 \\
Nephrology & 30 & 13.3 & 0.59 & 8.4 \\
Emergency medicine & 42 & 26.2 & 0.57 & 19.1 \\
Infectious diseases & 36 & 13.9 & 0.57 & 10.0 \\
Endocrinology & 51 & 11.8 & 0.52 & 4.0 \\
Neurology & 37 & 10.8 & 0.49 & 7.7 \\
\bottomrule
\end{tabular}

\end{table}

Emergency medicine is the outlier on both axes: its unanimous consensus is wrong about twice as often as the layer average, and it is the one specialty whose realized risk under the layer-level threshold exceeds the target. This is consistent with clinical experience, as acute multi-system presentations are where a shared quick prior is most likely to mislead all eight candidates at once. Coverage does not simply track the error rate: endocrinology and neurology have among the lowest unanimous-layer error rates yet receive the most conservative coverage, while obstetrics and cardiology combine similar error rates with far wider certified regions, meaning the process signals are differently informative across specialties, not merely calibrated to difficulty. The spread matters for deployment: a layer-level certificate is honest in aggregate, and a site that needs per-specialty control, routing emergency-medicine questions more conservatively than cardiology ones, can re-run the same calibration within specialty strata once enough per-specialty calibration data accumulate.

\section{Cost Accounting}\label{app:cost}
The passive layers are computed from logs the system already writes, at no inference cost. The probe fires only on unanimous first-round votes (\UaRateMedQA\% of MedQA, \UaRateMMLUPro\% of MMLU-Pro medicine, \UaRateMedMCQA\% of MedMCQA) and costs one counter-query call, one retrieval round, and $N$ concurrent re-answers: a median of \ProbeMedianSecMedQA{}\,s and \ProbeMedianTok{} completion tokens per probed MedQA question, against the \GenMedianSecMedQA{}\,s the substrate already spends answering, so probing roughly doubles the cost of a unanimous question and adds nothing to a contested one. Rationale scoring is a single batched NLI pass, minutes per dataset on one GPU. All experiments ran on one node with six A40 GPUs; no API models were used.

\section{Qualitative Cases}\label{app:qual}
The two cases of Figure~\ref{fig:motivation}, with their probe trajectories as logged. Both end in a unanimous panel; what differs is which answer survives.

\textbf{Case 1: unanimously wrong, repaired ($\keeprate = 0$).}
A patient hospitalized for a week after a stroke develops fever, purulent cough, and a new consolidation; all eight candidates answer \emph{Streptococcus pneumoniae}, each rationale invoking the same community-acquired prior and none engaging the hospitalization timeline. The reasoning is not careless: one rationale methodically rules out tuberculosis and \emph{Haemophilus influenzae} before concluding \emph{``Given the clinical presentation, the most likely causal microorganism is Streptococcus pneumoniae''}; each step is sound, and the error is in what the chain never considers. The probe's counter-query generator re-reads the stem and targets the discarded cue, generating \emph{``Hospital-acquired pneumonia in stroke patients: common pathogens''} and \emph{``Role of Staphylococcus aureus in hospital-acquired pneumonia with consolidation''} among its four queries. With the eight retrieved challenge documents in context, all eight re-answers abandon the consensus and converge, again unanimously, on the gold answer \emph{Staphylococcus aureus}. The probe did not surface unusual evidence; it directed attention back to context the candidates had ignored.

\textbf{Case 2: unanimously correct, held ($\keeprate = 1$).}
On the dimeric-immunoglobulin question, all eight candidates answer that secretory IgA inhibits bacterial adherence to mucous membranes, and the eight rationales cite the same mechanism. The probe generates genuine counter-queries, one asking directly for \emph{``Evidence against IgA's role in inhibiting bacterial adhesion''}, and returns eight challenge documents. All eight re-answers weigh the challenge and retain the consensus.

These two trajectories are the endpoints of the keep rate; intermediate values, a consensus that partially fragments under challenge, populate the separation of Figure~\ref{fig:flip}. The complete probe records, including all generated queries, retrieved document identifiers, and per-candidate re-answer votes, are in the released logs.

\section{Implementation Details}\label{app:impl}
The substrate runs unmodified at a pinned commit, and every component downstream of it is deterministic given the seed. Entailment is scored by DeBERTa-large fine-tuned on MNLI \citep{he2021deberta}; reranking uses the MedCPT cross-encoder \citep{jin2023medcpt}; both are used off the shelf without fine-tuning. Table~\ref{tab:hyper} lists the hyperparameters of each stage. None of them were tuned on evaluation data: the generation and retrieval settings are the substrate's own defaults, the classifier is a scikit-learn default configuration, and the certification constants of Appendix~\ref{app:cert} were fixed before calibration data were collected. A single seed governs generation, data splits, cross-validation, and bootstrap resampling; exact software versions and the per-dataset settings manifest ship with the code.

\begin{table}[h]
\centering
\caption{Hyperparameters by pipeline stage.}
\label{tab:hyper}
\small
\begin{tabular}{lc}
\toprule
Parameter & Value \\
\midrule
\multicolumn{2}{l}{\emph{Consensus generation}} \\
Panel size $N$ & $8$ \\
Maximum rounds $T$ & $8$ \\
Candidate temperature & $0.7$ \\
Candidate max new tokens & $2048$ \\
Query temperature (thinking mode) & $0$ \\
\addlinespace
\multicolumn{2}{l}{\emph{Retrieval}} \\
BM25 $(k_1, b)$ & $(0.9,\ 0.4)$ \\
Candidates per query & $32$ \\
Kept after reranking & $2$ \\
Documents per round & $\le 8$ \\
\addlinespace
\multicolumn{2}{l}{\emph{Rationale scoring and process score}} \\
NLI input limit (tokens) & $256$ \\
Classifier & $\ell_2$ logistic regression \\
Cross-validation folds & $5$ \\
\addlinespace
Random seed (all stages) & $42$ \\
\bottomrule
\end{tabular}
\end{table}

\end{document}